\documentclass[letterpaper, 10 pt, conference]{ieeeconf}  

\IEEEoverridecommandlockouts                              

\usepackage{amssymb,amsmath}
\usepackage{gensymb}
\usepackage{cite}
\usepackage{booktabs}

\newcommand{\sref}[1]{Section~\ref{#1}}
\newcommand{\figref}[1]{Fig.~\ref{#1}}
\newcommand{\tabref}[1]{Table~\ref{#1}}

\usepackage{hyperref}
\hypersetup{bookmarksopen,bookmarksnumbered,
pdfpagemode=UseOutlines,
colorlinks=true,
linkcolor=blue,
anchorcolor=blue,
citecolor=blue,
filecolor=blue,
menucolor=blue,
urlcolor=blue
}

\newcommand{\IN}{{\it in}}
\newcommand{\ON}{{\it on}}

\newcommand{\grasped}{$O_G$}
\newcommand{\target}{$O_T$}

\newcommand{\ego}[1]{$M^{\mathrm{ego}}_{#1}$}
\newcommand{\obj}[1]{$M^{\mathrm{obj}}_{#1}$}
\newcommand{\egoobj}[1]{$M^{\mathrm{ego+obj}}_{#1}$}
\newcommand{\egoPobj}[1]{$M^{\mathrm{ego+pre}}_{#1}$}

\newcommand{\zv}{$\textcolor{red}{\vec{0}}$}

\usepackage{soul}
\usepackage{xcolor}
\usepackage{colortbl}
\usepackage{pifont}
\newcommand{\cmark}{\ding{51}}%
\newcommand{\xmark}{\ding{55}}%
\usepackage{multirow}  
\usepackage[prependcaption,textsize=tiny]{todonotes}

\title{\LARGE \bf
Improving Robot Success Detection using Static Object Data 
}

\author{Rosario Scalise, Jesse Thomason, Yonatan Bisk, and Siddhartha Srinivasa%
\thanks{Paul G. Allen School of Computer Science and Engineering,
        University of Washington
        {\tt\small rosario@cs.washington.edu}}%
}

\begin{document}

\maketitle
\thispagestyle{empty}
\pagestyle{empty}

\begin{abstract}
We use static object data to improve success detection for stacking objects \ON{} and nesting objects \IN{} one another.
Such actions are necessary for certain robotics tasks, e.g., clearing a dining table or packing a warehouse bin.
However, using an RGB-D camera to detect success can be insufficient: same-colored objects can be difficult to differentiate, and reflective silverware cause noisy depth camera perception.
We show that adding static data about the objects themselves improves the performance of an end-to-end pipeline for classifying action outcomes.
Images of the objects, and language expressions describing them, encode prior geometry, shape, and size information that refine classification accuracy.
We collect over 13 hours of egocentric manipulation data for training a model to reason about whether a robot successfully placed unseen objects \IN{} or \ON{} one another.
The model achieves up to a 57\% absolute gain over the task baseline on pairs of previously unseen objects.
\end{abstract}

\section{Introduction~}
\label{sec:intro}

\begin{figure}[!t]
\centering
\includegraphics[width=\linewidth]{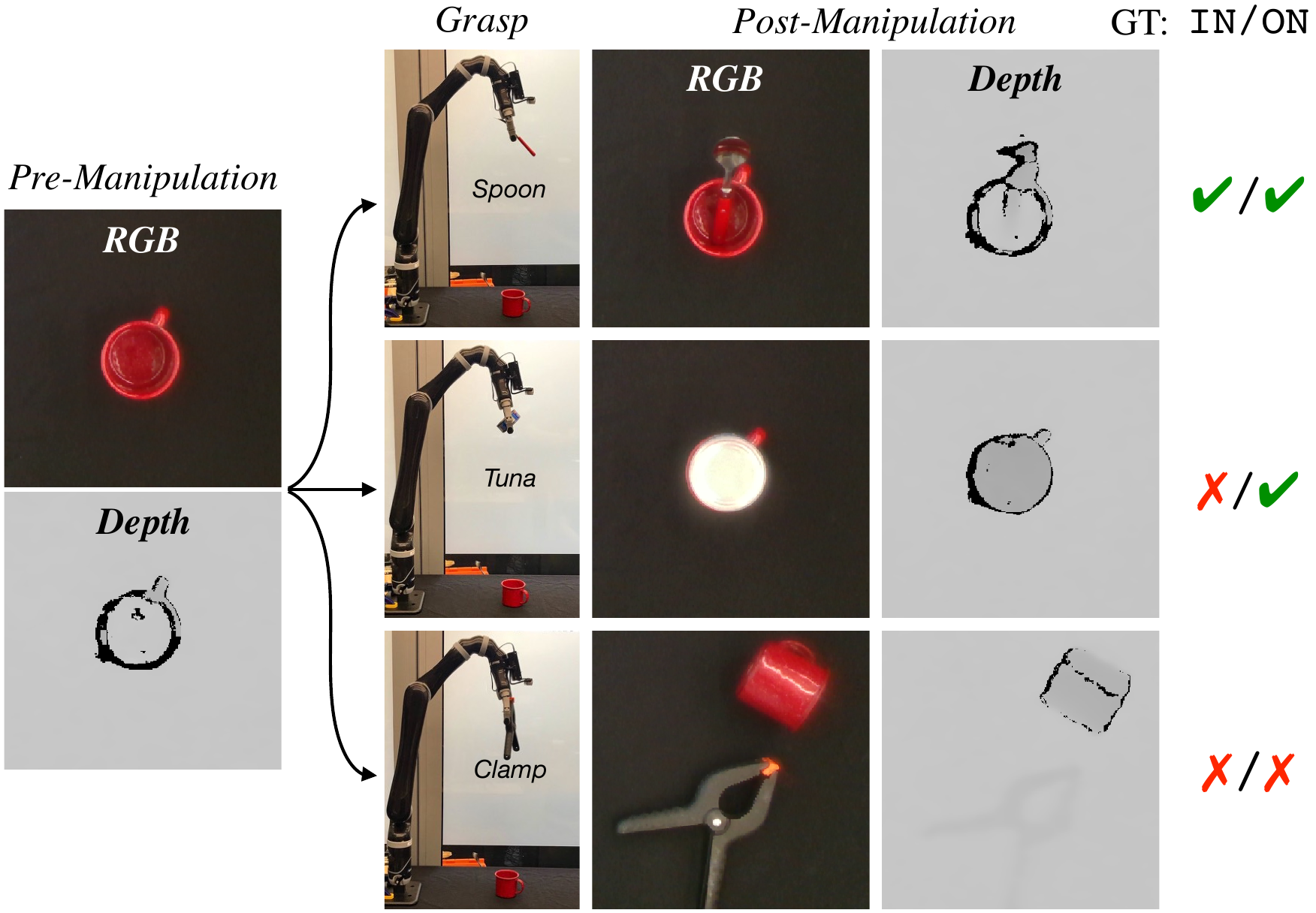}
\caption{The task is to classify whether dropping one object onto another resulted in the first being \IN{} or \ON{} the second.
Robot RGB-D scans of the workspace pre and post action are used these to predict the ground truth.}
\label{fig:best}
\end{figure}

We show that static vision and language data about the objects a robot manipulates can improve action outcome detection compared to a robot's egocentric camera data alone.
We consider the task of determining whether a robot's stacking action resulted in successfully placing an object \IN{} or \ON{} another object given egocentric scene information from a manipulator-mounted camera\footnote{\url{http://rosarioscalise.com/IROS19}} (\figref{fig:best}) and static vision and language data about the objects themselves in the form of images and referring expressions.

Consider a robot clearing a table of a plate, two cups, and two forks. 
Rather than clear each item one by one, a more efficient plan is to nest the cups \IN{} one another, place the forks \IN{} the cups, and put that collection \ON{} the plate.
Accomplishing this plan requires the robot to detect success after nesting and stacking actions.
Before the robot stacks the cups \ON{} the plate, it must know that it has successfully nested the cups \IN{} one another.

This problem is non-trivial for unseen objects with a wide range of geometries.
In the first row of \figref{fig:best}, the spoon is \IN{} the cup, but the robot's RGB signal is shallow because the handle and the cup are both red, and the depth signal is noisy because of the reflective surface of the spoon that hangs outside of the cup.

For such applications, the objects in a robot's workspace have properties---such as geometry, size, and shape---that provide a prior for action success.
These properties are reflected in static visual and language data about workspace objects.
Pictures taken from multiple viewpoints can reveal an object's geometry, while language descriptions can reveal an object's category (e.g., \emph{cup}) and shape (e.g., \emph{long}).

Our insight is to augment RGB-D workspace signals with vision and language priors about the objects themselves to improve classification accuracy of containment relationship after a robot stacking action on unseen objects.\footnote{All code and data for reproducing our results are available at\\ \url{https://github.com/thomason-jesse/YCBLanguage}}
In \figref{fig:best}, static images show that the cup has a hollow center, so objects can go \IN{} it, while language referring expressions (e.g., \emph{silver spoon with a long handle}) reveal a \emph{long} shape that is easy to rest \ON{} other objects.

\begin{figure*}[!t]
\centering
\includegraphics[width=\linewidth]{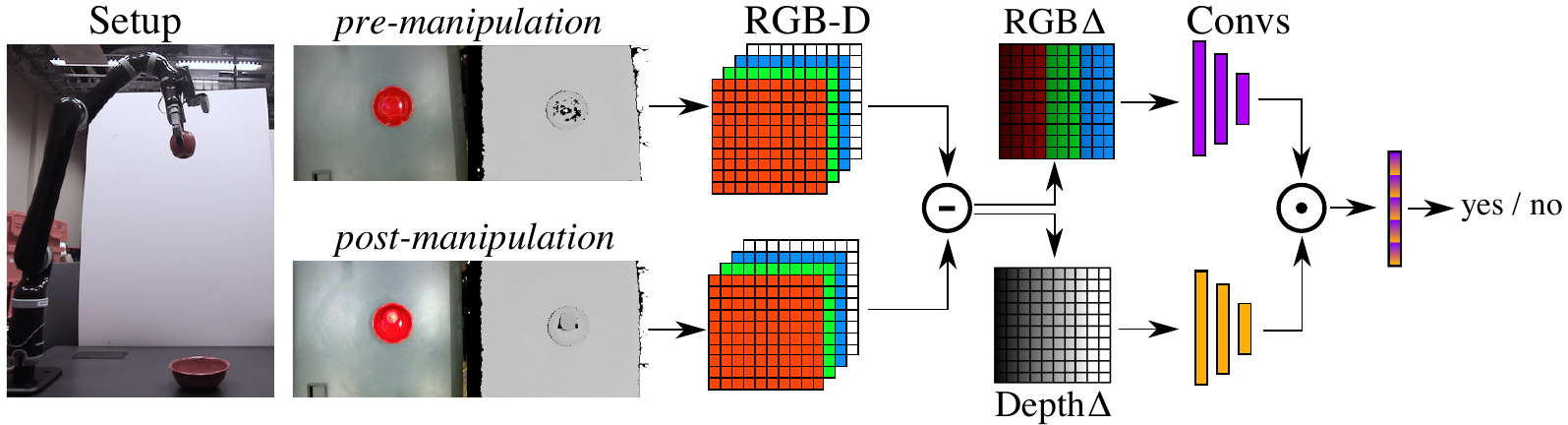}
\caption{End-to-end model \ego{} architecture trained to detect whether the grasped object is either \IN{} or \ON{} the target object after manipulation given egocentric scans.
The robot takes RGB-D snapshots of the workspace before and after attempting to stack the objects.
The color and depth differences between these scenes are processed by separate convolutional neural networks, then element-wise multiplied together and used to predict the outcome through a fully connected layer.
Both color and depth are informative for this task since objects can have similar colors and reflective surfaces can mislead depth scans.
}
\label{fig:robot_model}
\end{figure*}

We examine whether a grasped object (\grasped{}) lands \IN{} or \ON{} a target object (\target{}) when a robot performs a simple stacking action.
We gather egocentric RGB-D camera streams of stacking pairs of objects from the Yale-CMU-Berkeley (YCB) Object and Model set \cite{Calli:2015}, which reflect over 50 hours of robot operation by a researcher.
We propose end-to-end CNN-based models to detect whether object \grasped{} is \IN{} or \ON{} object \target{} after manipulation given this egocentric data (\figref{fig:robot_model}).
We improve performance by adding static object images and referring expressions.

\noindent \textbf{Contributions:} 
\begin{itemize}
\item We show that vision and language data about objects improves detection success for robot action outcomes;
\item we provide an end-to-end model for detecting successful placement of one object \IN{} or \ON{} another; and
\item we create a dataset of over 13 hours of robot stacking sensor streams, more than 4000 human annotations of whether object pairs can be \IN{} and \ON{} one another, and over 800 natural language object referring expressions.
\end{itemize}

We assume oracle-level object grasping (grasp selected by the researcher), and motion planning that results in object \grasped{} being held within~3cm of the target \target{} before being dropped.
We assume a fixed vantage point from which to survey the scene pre- and post-manipulation, ameliorating one difficulty of manipulator-mounted cameras.
We consider pairwise interactions between objects, limiting our task to two objects at a time.
We consider nesting objects \IN{} and stacking them \ON{} one another as a starting point for a wider class of object containment relations.

\section{Related Work~}
\label{sec:relatedwork}

Our task is to detect action success in terms of physical relationships between two objects using a manipulator-mounted camera.
This task relates to work in object relations, grasping and stacking, and egocentric vision.
We demonstrate improvements by using static data about the objects involved, including human annotations, related to other work in utilizing human judgements for robotics applications.

\paragraph{Object Relations}
Beyond a functional understanding of which object it can interact with~\cite{Ye2017WhatCI}, a robot can reason about relationships between objects.
Single models that take in information about two objects can reason about multiple relationships between them~\cite{santoro:nips17}, but this requires large, annotated training data for all inter-object relations.
Some work targets relationships between \emph{tools} and target objects~\cite{zhu2015understanding}.
Other work considers \emph{stacking} to be a category of high-level activity within a dataset~\cite{koppula2013learning}, or learns object-level attributes for \emph{stackable} and \emph{containment}~\cite{aldoma2012supervised}.
Still others learn to ground multi-object concepts, like \emph{rows} and \emph{columns}~\cite{paul:ijrr18}.
We use static object information as a prior to improve egocentric detection of whether a robot successfully stacked one object \ON{} or contained one object \IN{} another.

\paragraph{Grasping and Stacking}
Current robot systems are increasingly proficient at robustly grasping objects~\cite{zeng2018icra,mahler2019learning,tobin2018domain} and subsequently dropping, placing, and balancing them~\cite{Paolini2014ADS,jiang2012learning}.
The outcomes of these actions vary based on the dexterity of the robot manipulator~\cite{holladay2013object}.
Longer-horizon tasks, like stacking multiple objects, require awareness of intermediate action success.
Research has addressed: reasoning about visual stability of objects in simulation~\cite{Li2017VisualSP,Groth2018ShapeStacksLV,hamrick2011internal} and in a robot workspace~\cite{furrer2017autonomous}; detecting intermediate task success~\cite{worcester2014distributed,xue1995error}; and predicting whether a task will succeed before executing it~\cite{Nguyen2014AutonomouslyLT}.
We define success in terms of two object relationships: \IN{} and \ON{}, and we assume oracle-level grasping and localization during action execution.

\paragraph{Egocentric Vision}
Existing work detects action outcomes using extrinsic camera viewpoints~\cite{dogar2015multi}.
However, as robots move from research labs to homes, static cameras become impractical and intrusive.
Instead, egocentric cameras can be mounted to the robot chassis~\cite{tremblay2018deep,stentz2015chimp} or manipulator~\cite{papanikolopoulos1993visual,klingensmith2016articulated}.
Most egocentric vision research centers on identifying human~\cite{duckworth:ai19,Ryoo2015RobotCentricAP} and robot activities~\cite{Li2016WhatAI} or ego-driven scene prediction~\cite{jayaraman:iccv15}.
These perspectives are often obtained through head- or body-mounted cameras~\cite{Damen2018ScalingEV,Luo2017SceneSR,Ma2016GoingDI}.
We use a manipulator-mounted camera, which may become the norm for future robot platforms (e.g., Kinova Jaco 3).
This setup restricts the robot to a single viewpoint at a time.

\paragraph{Human Judgement}
Attempting to model or utilize human manipulation abilities is a long-studied problem in robotics~\cite{iberall:icra88}.
Researchers have used crowdsourcing platforms to source data for grasping novel objects \cite{Sorokin2010PeopleHR} and translating natural language commands to actions~\cite{bisk18,thomason:ijcai15}.
Natural language descriptions of objects from humans can help robots learn visual (\emph{red})~\cite{yang:corl18,tucker:isrr17,parde:ijcai15} or multi-modal (\emph{heavy})~\cite{thomason:corl17,thomason:ijcai16} classifiers between words and workspace objects, as well as inter-object relations~\cite{kulick:ijcai13}.
While embeddings learned by language modeling (e.g.,~\cite{pennington:emnlp14}) place words like \emph{fork}, \emph{plate}, and \emph{cup} as neighbors in the embedding space despite their referents having different physical properties, language vectors informed by physical properties~\cite{thomason:ijcai16} encode some differences in geometry.
However, such vocabularies are noisy and sparse, making them insufficient for an open-vocabulary (unrestricted word choice) domain.
We crowdsource open-vocabulary referring expressions (e.g., ``yellow long fruit'') for objects, and annotations for whether humans think objects can be placed \IN{} or \ON{} one another.

\section{Task and Data~}
\label{sec:task_and_data}

Our task is to determine whether a robot stacking action successfully placed object \grasped{} \IN{} or \ON{} object \target{} given egocentric scans from a manipulator-mounted camera.

\paragraph{Data Folds}
We chose the YCB Object and Model set \cite{Calli:2015}, a standardized collection of physical objects designed for benchmarking robot manipulation, for two reasons.
First, this object set is widely used among robotics labs, making the data we collect useful to other researchers.
Second, its constituent objects span a diverse space of geometries and potential physical interactions, making the detection of simple physical relations like \IN{} and \ON{} after a stacking attempt more challenging.

Among the full set of 90 YCB objects, $Y$, we designate a subset of 28 \textit{containers} as $C$ (e.g., cups, boxes, and cans).
We split $Y$ into \emph{Train}, \emph{Development}, and \emph{Test} data folds.
Pairs of objects considered in each fold are considered for the task (e.g., all \emph{Train} object pairs (\grasped{}, \target{}) comprise two \emph{Train} objects).
Consequently, not all object pairs are included in the data folds since pair $i, j$ cannot be included in any fold if individual objects $i$ and $j$ are assigned to different folds.
This split also means that all objects are unseen at inference time.
For testing \grasped{} \IN{} \target{}, we consider pairs where $($\grasped$, $\target$)\in Y\times~C$, all objects paired with all containers; for \grasped{} \ON{} \target{}, 
we consider pairs where $($\grasped$, $\target$)\in Y\times~Y$, all objects paired with all objects.
We evaluate on the subset of object pairs in those splits for which we collect egocentric, RGB-D data from robot trials ({\bf Robot Pairs}), and we define an auxiliary task over all pairs in these splits ({\bf All Pairs}).
\tabref{tab:folds_summary} gives the number of pairs per data fold.

\paragraph{Robot Pairs}
\label{sec:robot_trials}
For a subset of {\bf All Pairs}, we collect RGB-D directly from the robot's egocentric viewpoint.
For each such {\bf Robot Pair}, five trials were gathered, resulting in $191\times 5=955$ training examples.
Our robot is the widely used Kinova Jaco$2$\cite{KinovaJ2}.
We use an Intel Realsense D415 RGB-D camera\cite{intel} mounted on the end-effector.

To collect egocentric camera data for each pair (\grasped{},~\target{}), object \target{} is first set on an origin position on the table.
Then, the robot moves to a designated survey position and captures a \emph{pre-manipulation} RGB-D frame.
An operator manually places object \grasped{} within the robot's gripper, and the gripper is closed.
Then, the robot gripper moves to a height of 3cm above \target{} and opens.
Finally, the robot returns to the survey position and captures a \emph{post-manipulation} RGB-D frame.

We collect 5 such trials per object pair. For each pair, the operator provides an annotation in \{\emph{Yes}, \emph{No}\} of the outcome, noting whether object \grasped{} was \IN{} or \ON{} object \target{} for all five trials.\footnote{For cases where the trials had different annotated outcomes (e.g., \ON{} 2 times and not 3 times), the label was decided as \emph{No}.
Similarly, annotator disagreement from Mechanical Turk was rounded to \emph{No}.}
This data collection is expensive to conduct because it uses operator-selected oracle grasping.

\begin{table}[t]
\caption{Dataset folds sizes.}
\begin{center}
\begin{small}
\centering
\begin{tabular}{l r r r r r r}
    \toprule
    & \multicolumn{2}{c}{\bf Objects} & \multicolumn{2}{c}{\bf Robot Pairs} & \multicolumn{2}{c}{\bf All Pairs} \\
	\bf Fold & $Y$ & $C$ & \IN{} & \ON{} & \IN{} & \ON{} \\
	\midrule
	Train & $51$ & $17$ & $191$ & $191$ & $800$ & $2500$ \\
	Dev & $20$ & $5$ & $47$ & $58$ & $100$ & $400$ \\
	Test & $19$ & $6$ & $60$ & $60$ & $114$ & $361$ \\
	\bottomrule
\end{tabular}
\end{small}
\end{center}
\label{tab:folds_summary}
\end{table}

\paragraph{Data Augmentation}
\label{sec:data_augmentation}
To augment expensive robot pair data collection, we gather complementary data about objects and object pairs via crowdsourcing through Amazon Mechanical Turk (AMT)\footnote{\url{https://www.mturk.com/}} conditioned only on static images of the objects.
The YCB dataset provides standardized images of each object from multiple angles.
For each object, we choose the 0$\degree$ yaw, 30$\degree$ pitch viewpoint as a representative image to show AMT workers (examples in Figure~\ref{fig:joint_model}).

Each AMT worker annotates 35 pairs of side-by-side object images.
For 9 of these pairs, we ask annotators whether object \grasped{} could be put \IN{} object \target{}.
For the remaining 26 pairs, we ask annotators if object \grasped{} could be placed \ON{} object \target{}. 
For each object pair, we collect at least three annotations for each of \IN{} and \ON{} and assign labels \{\emph{Yes}, \emph{No}\} based on them.

Subsequently, for each YCB object from $Y$, we collect three referring expressions from three different annotators (nine in total per object).
These expressions, for example \emph{small black sphere} in the first row of \tabref{tab:cherry_lemon}, provide linguistic clues about object size and shape properties.

\section{Models~}
\label{sec:models}

\begin{figure*}
    \centering
    \includegraphics[width=\linewidth]{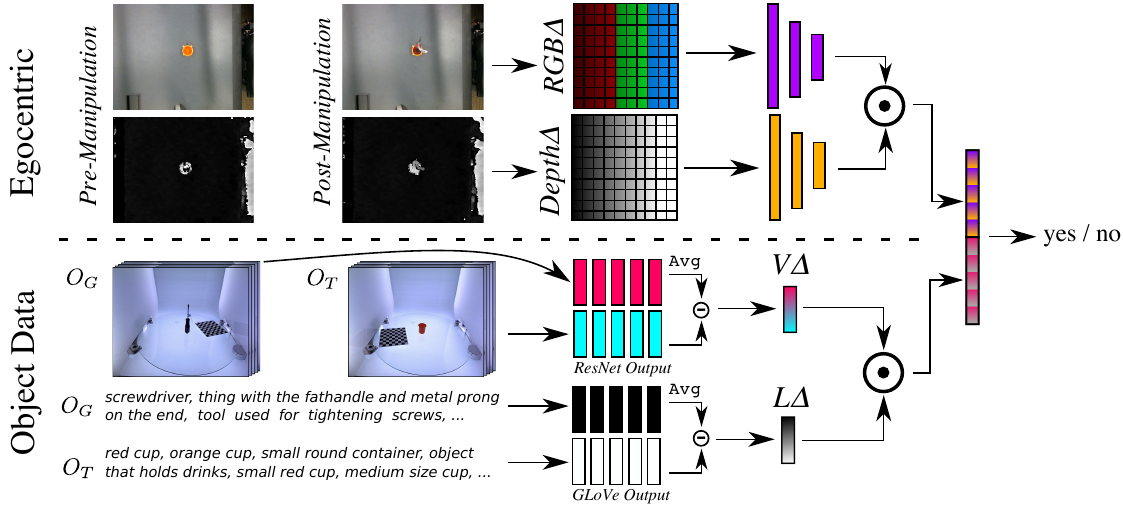}
    \caption{
    End-to-end model \egoobj{} takes egocentric visual input, vision embeddings from multiple static image viewpoints of each object, and language embeddings from referring expressions for each object.
    The vision and language embeddings $\Delta$s for each object pass through fully connected layers, whose outputs are element-wise multiplied and then concatenated with the output of the egocentric pipeline.
    This concatenation is used to classify the pair label.
    }
    \label{fig:joint_model}
\end{figure*}

We introduce an end-to-end model that uses egocentric camera inputs to determine whether a stacking action resulted in an \IN{} or \ON{} relationship between two objects (\figref{fig:robot_model}).
We then add additional object data inputs: images and natural language referring expressions of the objects.

\paragraph{Egocentric Model \ego{}}
\label{sec:robot_model}
During a robot execution trial, the robot's egocentric camera provides \emph{pre-manipulation} and \emph{post-manipulation} RGB-D inputs: $R_{pre}$ and $R_{post}$ of size $640\times 480$.
These images are downsampled once via an average pooling layer to size $64\times 48$. We use the difference between the images ($R_{\Delta}$) as input to the model.

We learn two egocentric classification models, \ego{in} and \ego{on}, that independently learn whether \IN{} and \ON{} hold between pairs of objects.
These models consist of two 3-layer convolutional nets, one for processing RGB and one for processing D inputs, with max-pooling followed by a singular fully connected layer.
Specifically, these models take $48 \times 64$ RGB (3 color channels + 1 $L_2$ channel) and depth image (1 depth channel) differences as input and output a label $c\in\{Yes, No\}$. To capture changes in RGB space, we include the $L_2$ distance between $R_{post}$ and $R_{pre}$.
$L_2$ lets us capture change across all three dimensions, while subtraction assumes the color channels are independent.
\figref{fig:robot_model} depicts the egocentric model architecture.

\paragraph{Egocentric plus Object Data Model \egoobj{}}
\label{sec:joint_model}
We augment the \emph{Egocentric Model} with two additional inputs.
First, we add five images of each object (\grasped{} and \target{}), independently representing front, back, left, right, and top-down views.\footnote{Extracted from given YCB object metadata. Accuracy is slightly lower when considering only the front image.}
These static images provide multiple vantage points from which to extract the objects' geometry features.
Second, we add natural language referring expressions for each object, which provide linguistic clues about physical properties (e.g., ``small'' and ``long'').

We use ResNet-152~\cite{he:arxiv15} and GLoVe vectors~\cite{pennington:emnlp14} (sans fine-tuning) to embed our images\footnote{Such models could potentially extract features from egocentric scans, but are trained for ImageNet classification, not task-relevant spatial relations.} and language.  
The embeddings across the five images and across the referring expression words are averaged to compute embeddings for the auxiliary vision and language input, respectively.
We compute the differences between embeddings for \grasped{} and \target{} visually and linguistically before projecting them to the same size space and concatenating the output to the penultimate layer of the \emph{Egocentric Model} for classification.
These object pair embedding vectors from egocentric scans and from object data concatenated for classification have the same size.
\figref{fig:joint_model} shows the model architecture.
We again learn independent \egoobj{in} and \egoobj{on} models.

Object images can capture geometry, e.g., if \target{} is wide and flat, \grasped{} is more likely to go \ON{} it.
Referring expressions can reveal how descriptor words like ``small'' and ``large'' for \grasped{} encode how likely it is to go \IN{} \target{}.

\paragraph{Egocentric plus Pretrained Object Data Model \egoPobj{}}
\label{sec:pretrained_model}
Running the robot through thousands of trials and labeling the outcomes is prohibitively expensive, but object data can be crowdsourced (as described in \sref{sec:task_and_data}).
Whereas in the previous two models, we trained using the smaller data subset, {\bf Robot Pairs}, here we make use of the {\bf All Pairs} data to pretrain relevant layers of the \egoobj{} architecture to obtain the \egoPobj{} model.

We pretrain the \egoobj{in} and \egoobj{on} architecture layers that map from input embeddings to hidden projection layers on an auxiliary task:
predicting \IN{} and \ON{} Mechanical Turk annotations given only images and referring expressions for objects \grasped{} and \target{} (e.g., no RGB-D input information).
This is a noisy signal because human intuition about these stacks might not align with the robot's ability, but we can train on the much larger set of {\bf All Pairs}.

We hypothesize that this pretraining will lead to higher task accuracy.
We denote \egoPobj{in} and \egoPobj{on} models with layers for processing image and referring expression inputs instantiated with pretrained weights from this auxiliary task.
We compare these against the randomly initializing the weights before training, as done for the \egoobj{} models.

\section{Experiments and Results~}
\label{sec:results}

We evaluate these models by comparing their performance on the \emph{Test} data fold of {\bf Robot Pairs}.\footnote{We tune all architecture hyperparameters to maximize accuracy on the development data fold. We use an Adam optimizer~\cite{kingma:iclr15} with learning rate $0.01$, dropout of $0.3$ and ReLu connections between fully connected layers, and a hidden layer size of $64$. For RGB-D, we use a $3\times 3$ kernel, with three layers of alternating convolution and max pooling operations and filters doubling at each.}
We compare model variants against one another and majority class and random baselines.
For evaluation, we train each model for 30 epochs, recording the average maximum test data fold accuracy reached in that time across 10 different seeds.\footnote{Random restarts to expose variance in model performance.}
During training, all 5 trials per object pair serve as individual training examples, while at inference time a majority vote across the 5 trials of the test pair is taken to assign the label.
\tabref{tab:results} shows model and baseline accuracy and standard deviation on the \emph{Development} and \emph{Test} data folds.

\begin{table}[t]
\caption{Model performance on both \emph{Development} and \emph{Test} {\bf Robot Pairs}.}
\centering
\begin{tabular}{l c c r r}
	\toprule
    & \multicolumn{2}{c}{\bf Model ($M$)} & \multicolumn{2}{c}{\bf Detection Correct $\uparrow$} \\
	&  Ego &  Object & \multicolumn{1}{c}{\IN{}} & \multicolumn{1}{c}{\ON{}} \\
	\midrule
	\multirow{5}{*}{\rotatebox{90}{\emph{Dev} Fold}}
	& \checkmark &          & $.69\pm.08$ & $.55\pm.11$ \\
	& \checkmark & \checkmark & $.70\pm.09$ & $.59\pm.07$ \\
	& \checkmark & pre        & $.73\pm.09$ & $.62\pm.07$ \\
	\cmidrule{2-5}
	& \multicolumn{2}{l}{Baseline (MC)} & $.32\pm.00$ & $.36\pm.00$ \\
	& \multicolumn{2}{l}{Baseline (Rand)} & $.49\pm.06$ & $.50\pm.06$ \\
	\midrule
	\multirow{5}{*}{\rotatebox{90}{\emph{Test} Fold}} 
	& \checkmark &            & $.77\pm.05$ & $.53\pm.10$ \\
	& \checkmark & \checkmark & $.74\pm.07$ & $.59\pm.08$ \\
	& \checkmark & pre        & $.77\pm.05$ & $.59\pm.06$ \\
	\cmidrule{2-5}
    & \multicolumn{2}{l}{Baseline (MC)} & $.20\pm.00$ & $.32\pm.00$ \\
    & \multicolumn{2}{l}{Baseline (Rand)} & $.52\pm.05$ & $.51\pm.07$ \\
	\bottomrule
	\multicolumn{5}{p{5cm}}{\checkmark indicates signal was included, while ``pre" indicates models with object features pretrained from {\bf All Pairs} data.}
\end{tabular}
\label{tab:results}
\end{table}

\begin{table*}[ht!]
    \caption{
    Egocentric \ego{} model and augmented \egoPobj{} model predictions. We display only the post-manipulation frame for $R_\Delta$.
    }
    \centering
    \footnotesize
    \begin{tabular}{@{}c>{\columncolor[gray]{0.9}}ccp{16em}>{\columncolor[gray]{0.9}}ccp{9em}@{}}
    \toprule
    \bf Egocentric
    \hfill $\Rightarrow$ & Pred & 
    \multicolumn{2}{l}{\bf Egocentric + Pretrained Object} \hfill $\Rightarrow$ & Pred & Truth & Reason\\
    \midrule
    \raisebox{-.8\height}{\includegraphics[width=0.15\linewidth]{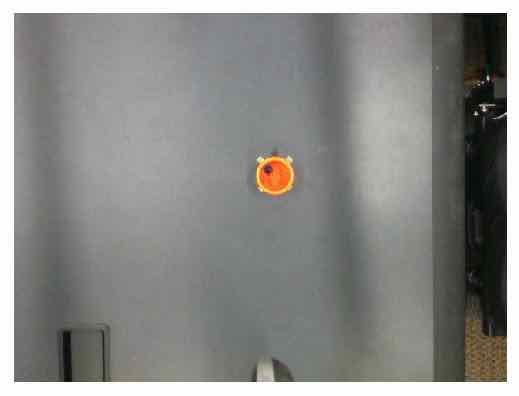}} 
    & \multirow{2}{*}{\cellcolor{red!25}\xmark In} 
        & \raisebox{-0.8\height}{\includegraphics[width=0.14\linewidth]{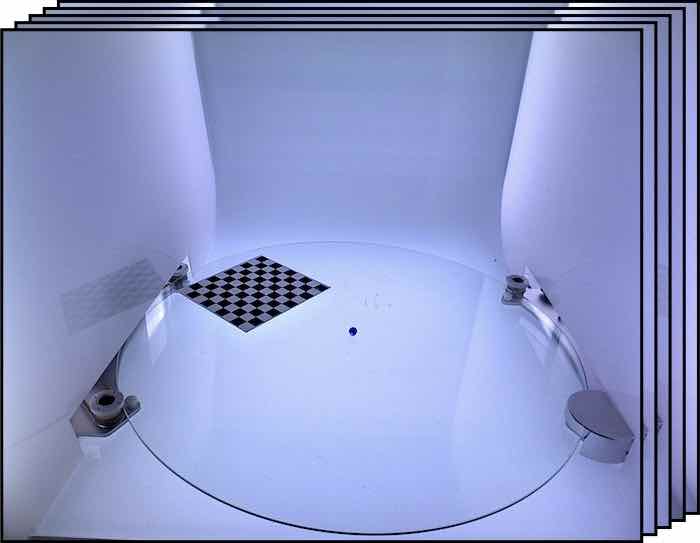}} & \emph{small black sphere}, \emph{round black item}, \emph{small marble}, \emph{the blue object}, \emph{round object}, \emph{tiny object}, \emph{tiny dot}, \emph{blue round object}, \emph{little ball} & \multirow{2}{*}{\cellcolor{blue!25} \cmark In} 
        & \multirow{2}{*}{\cmark In} & \multirow{2}{9em}{With egocentric vision alone, it is difficult to see the small marble.
        After adding object data, the model classifies the \emph{tiny} marble \IN{} the \emph{cup} container.} \\
    \raisebox{-.8\height}{\includegraphics[width=0.15\linewidth]{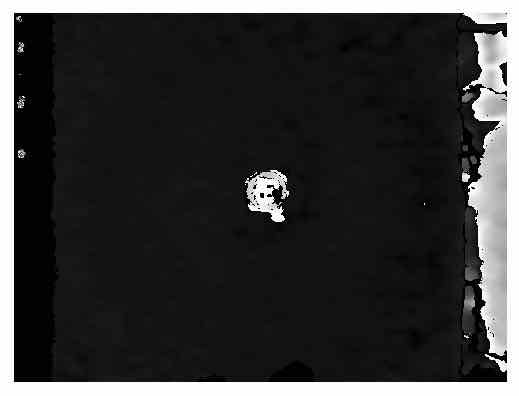}}
    & \cellcolor{red!25}
        & \raisebox{-.8\height}{\includegraphics[width=0.14\linewidth]{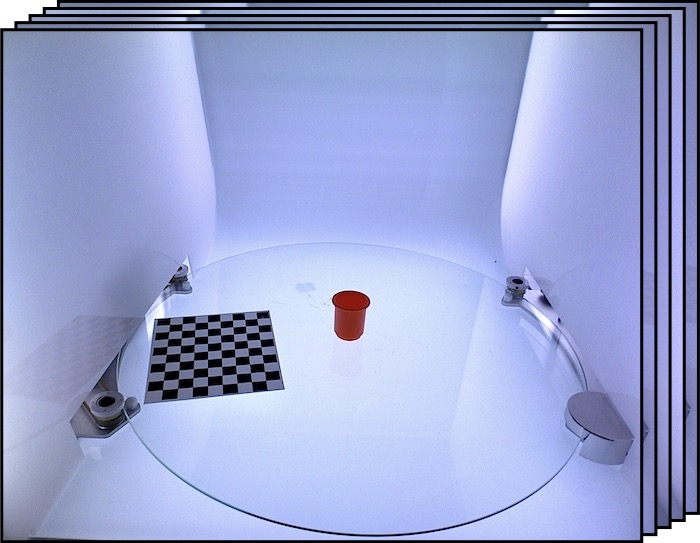}} & \emph{red cup}, \emph{orange cup}, \emph{small round container}, \emph{object that holds drinks}, \emph{small red cup}, \emph{red cup}, \emph{medium size cup without handles}, \emph{red plastic thing}, \emph{red cylinder} & \cellcolor{blue!25} & &  \\
    \midrule
       \raisebox{-.9\height}{\includegraphics[width=0.15\linewidth]{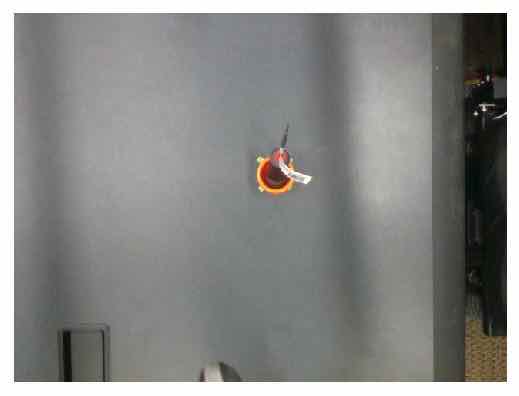}} 
       & \multirow{2}{*}{\cellcolor{red!25} \xmark In} 
        & \raisebox{-.9\height}{\includegraphics[width=0.14\linewidth]{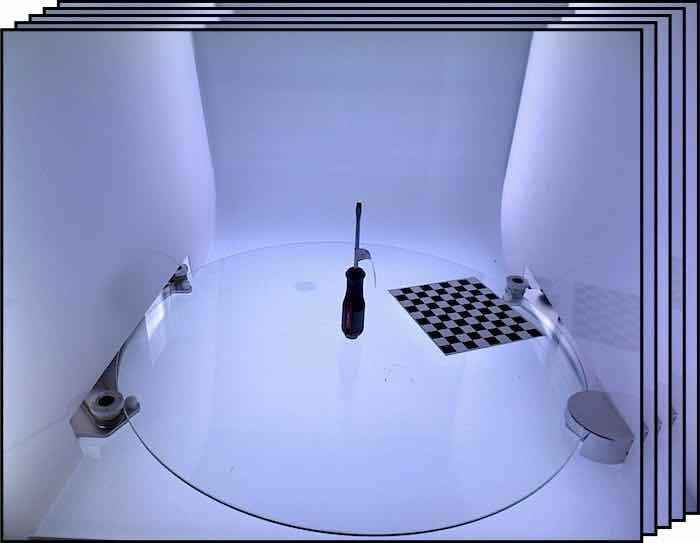}} & \emph{screwdriver}, \emph{thing with the fat handle and metal prong on the end}, \emph{tool used for tightening screws}, \emph{screw driver with long tip}, \emph{screwdriver}, \emph{plastic handle screw driver}, \emph{non phillips screw driver}, \emph{tool}, \emph{black screwdriver} & \multirow{2}{*}{\cellcolor{blue!25} \cmark In} 
        & \multirow{2}{*}{\cmark In} & \multirow{2}{9em}{The reflective screwdriver and its tag spill over the edge of the cup, creating a misleading egocentric depth scan.
        After adding object data, the model classifies the \emph{screwdriver} as being \IN{} the \emph{cup} container.} \\
    \raisebox{-.8\height}{\includegraphics[width=0.15\linewidth]{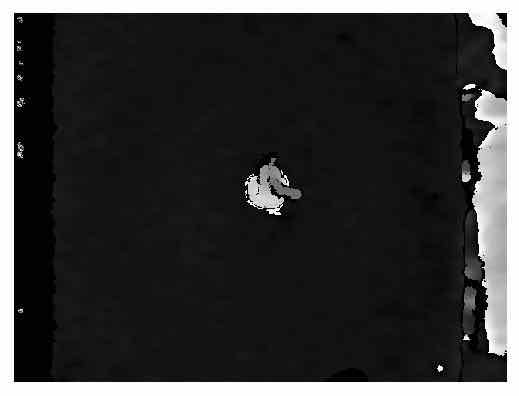}} 
    & \cellcolor{red!25}
        & \raisebox{-.8\height}{\includegraphics[width=0.14\linewidth]{figs/065-a_cups-stack.jpeg}} & \emph{red cup}, \emph{orange cup}, \emph{small round container}, \emph{object that holds drinks}, \emph{small red cup}, \emph{red cup}, \emph{medium size cup without handles}, \emph{red plastic thing}, \emph{red cylinder} & \cellcolor{blue!25} &  &  \\
    \midrule
       \raisebox{-.8\height}{\includegraphics[width=0.15\linewidth]{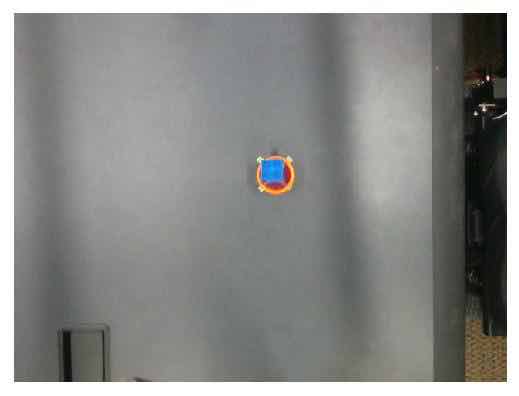}} 
       & \multirow{2}{*}{\cellcolor{red!25} \xmark On} 
        & \raisebox{-.8\height}{\includegraphics[width=0.14\linewidth]{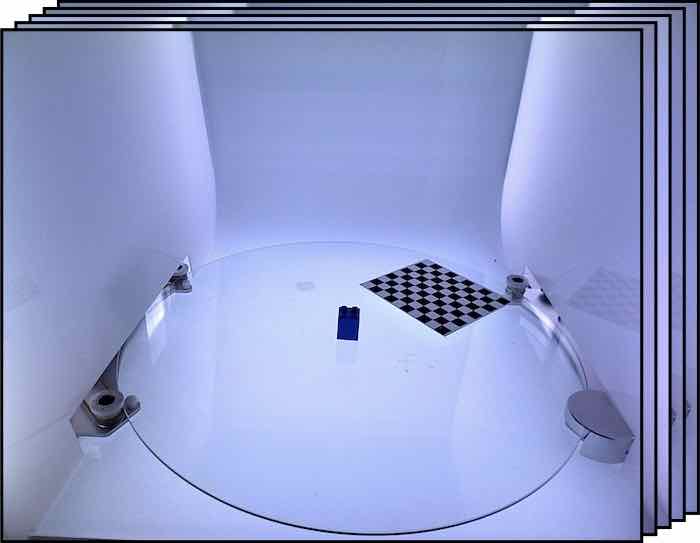}} & \emph{blue thing}, \emph{blue plastic rectangle}, \emph{blue plastic block}, \emph{blue cube}, \emph{lego piece}, \emph{blue plastic thing}, \emph{blue block}, \emph{small square block}, \emph{little blue block} & \multirow{2}{*}{\cellcolor{blue!25} \cmark On} 
        & \multirow{2}{*}{\cmark On} & \multirow{2}{9em}{The block rests partially nested in the cup, exhibiting a rare hybrid of both \ON{} modalities (resting on top versus nested inside).
        After adding object data, the \emph{small block} is classified as resting \ON{} the \emph{cup} container.} \\
    \raisebox{-.8\height}{\includegraphics[width=0.15\linewidth]{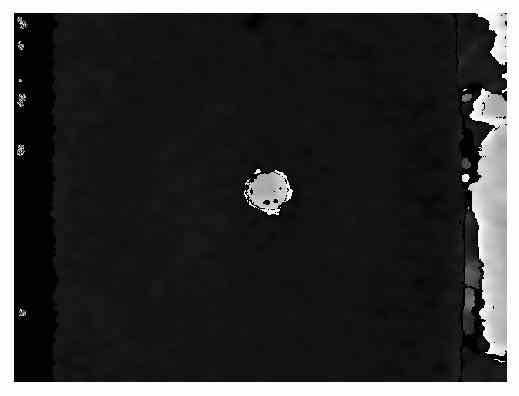}} 
    & \cellcolor{red!25}
        & \raisebox{-.8\height}{\includegraphics[width=0.14\linewidth]{figs/065-a_cups-stack.jpeg}} & \emph{red cup}, \emph{orange cup}, \emph{small round container}, \emph{object that holds drinks}, \emph{small red cup}, \emph{red cup}, \emph{medium size cup without handles}, \emph{red plastic thing}, \emph{red cylinder} & \cellcolor{blue!25} & &  \\
    \midrule
    \raisebox{-.9\height}{\includegraphics[width=0.15\linewidth]{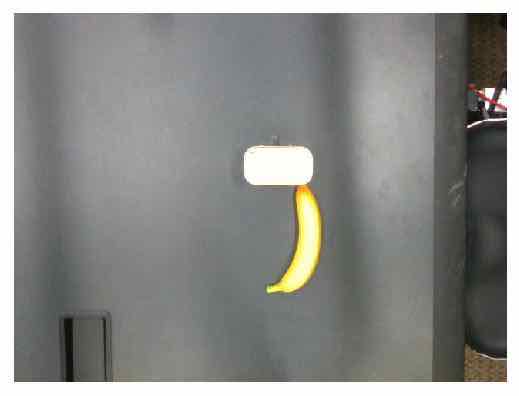}} 
    & \multirow{2}{*}{ \cellcolor{blue!25} \xmark On} 
        & \raisebox{-0.9\height}{\includegraphics[width=0.14\linewidth]{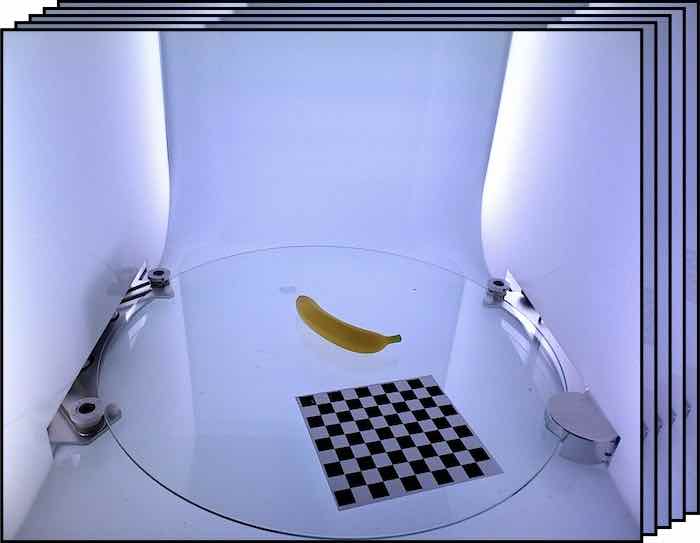}} & \emph{yellow thing}, \emph{long yellow item}, \emph{soft yellow thing}, \emph{yellow curved cylinder}, \emph{yellow fruit}, \emph{the object that is mostly yellow with slight green at one of the tips}, \emph{yellow long fruit}, \emph{yellow banana}, \emph{banana} & \multirow{2}{*}{\cellcolor{red!25} \cmark On} 
        & \multirow{2}{*}{\xmark On} & \multirow{2}{9em}{Given object data predictions pretrained on crowdsourced labels, the model predicts that \emph{fruit} is balanced \ON{} the \emph{can}, consistent with human ability.
        In reality, the robot manipulator lacks the dexterity to reliably balance the banana.} \\
    \raisebox{-.8\height}{\includegraphics[width=0.15\linewidth]{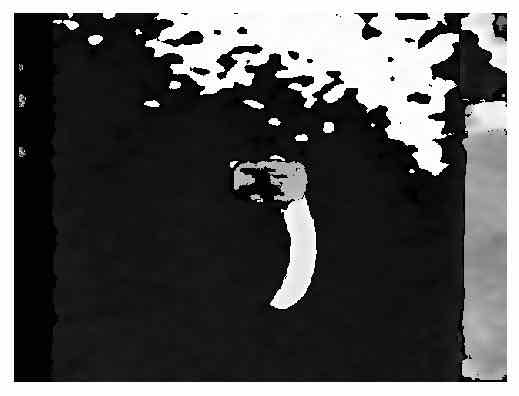}}
    & \cellcolor{blue!25}
        & \raisebox{-.8\height}{\includegraphics[width=0.14\linewidth]{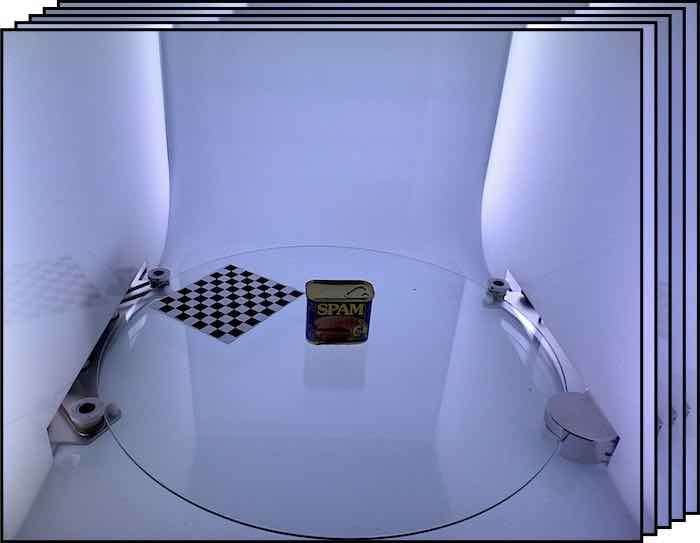}} & \emph{spam}, \emph{canned meat}, \emph{metal can}, \emph{can of spam}, \emph{aluminum cube}, \emph{blue and gold cube}, \emph{rectangular can}, \emph{square}, \emph{glass circle} & \cellcolor{red!25} & &  \\
    \bottomrule
    \end{tabular}
    \label{tab:cherry_lemon}
\end{table*}

All end-to-end models achieve higher accuracy than the baselines, with a 57\% absolute gain for the \IN{} task between the baseline and \ego{in} / \egoPobj{in} models.
When considering the \IN{} and \ON{} detection tasks together, models with vision and language processing layers pretrained on an auxiliary task, \egoPobj{}, score higher accuracy than the others across both the \emph{Test} and \emph{Dev} folds.\footnote{We also tested models pretrained on the auxiliary task but using only object data (e.g., no \emph{Ego} input). These are accurate for detecting the more geometry-informed \IN{} relation but fall short when classifying \ON{}, which is conditioned on robot dexterity (e.g., \tabref{tab:cherry_lemon} row 4).}

\section{Discussion~}
\label{sec:analysis}

The proposed models all outperform the baseline at detecting both \IN{} and \ON{} relationships.
Here, we discuss their relative performance on these tasks, and \tabref{tab:cherry_lemon} shows examples of model disagreements between the egocentric \ego{} models and object data augmented \egoPobj{} models.\footnote{\tabref{tab:cherry_lemon} shows the best performing model across random restarts.}

\paragraph{Analysis}

The egocentric RGB-D scans are not always sufficient to make an accurate success judgement.
The first and second rows of \tabref{tab:cherry_lemon} show examples where egocentric data fails to discern that object \grasped{} has been successfully placed in object \target{}.
In the first row, object \grasped{} is a tiny, barely perceptible marble.
In the second row, object \grasped{} is a screwdriver with a shiny tip, resulting in a misleading depth scan (similar to the first row of \figref{fig:best}).

In general, our models have more limited accuracy when detecting \ON{} than \IN{} relationships: the \ON{} relationship is semantically more general than \IN{}.
Specifically, consider that any object \grasped{} \IN{} a container \target{} is also implicitly \ON{} that container, while the converse does not hold.
This leads to at least two distinct modalities for \ON{}: object \grasped{} on top of object \target{}, and \grasped{} is inside of \target{}. 
Each presents differently in depth scans.
The third row of \tabref{tab:cherry_lemon} shows an object pair exhibiting a hybrid of these two \ON{} modalities.
Generalizing from the \emph{Train} to \emph{Test} data pairs is challenging, especially for the wider \ON{} relation.\footnote{By contrast, on the \emph{Train} data fold, all models (except the baselines) for both \IN{} and \ON{} perform at or above 85\% accuracy when measured at the epoch achieving the best \emph{Test}/\emph{Dev} generalization accuracy reported in \tabref{tab:results}.}

Human intuitions of what relationships can hold between objects are not perfectly consistent with the robot's capabilities.
The fourth row of \tabref{tab:cherry_lemon} shows that while humans annotate that a banana can balance \ON{} a spam can, the robot manipulator lacks the dexterity to reliably balance the cylindrical banana.
This is reflected in the egocentric \ego{on} model making the correct decision, while the object data augmented model is mislead by pretraining from human judgement that fruits can be balanced on cans.

\paragraph{Limitations and Future Work}
To explore the containment detection task, we make several data collection and modeling assumptions (listed in \sref{sec:intro}).
Relaxing these assumptions is beyond the scope of this work, but we discuss future research towards doing so.

Collecting egocentric camera stream information from a robot performing manipulation tasks is expensive in terms of operator hours, but running data collection in a self-supervised fashion is difficult in practice when dealing with unseen objects.
Future work towards automatically selecting robust grasps, and planning for correct object placement on the target object, may allow the robot to gather egocentric data autonomously for subsequent experimenter annotation.

We simplify the manipulator-mounted vision input to our model by assuming a fixed vantage point to capture pre- and post-manipulation RGB-D scans.
While our data processing makes the model robust to different table heights between pair samples (because we subtract pre- and post- scans), it may be brittle against lighting and axial changes.
Standard computer vision techniques (e.g., cropping, translating, color shifting) can increase model robustness, while future autonomous data collection could use the manipulator-mounted camera to capture scans from multiple vantage points.

We consider pairwise interactions between objects when both pre- and post-manipulation scans are available.
In many applications contexts (e.g., table clearing), multiple objects interact, and some may already be interacting when the robot arrives.
After nesting two cups \IN{} one another, a robot may need to determine whether they can rest \ON{} a plate.
Similarly, a cup with a fork already \IN{} it cannot have a cup nested \IN{} it.
Static object data generalizes to the multiple object domain in some cases.
Using referring expressions, we can infer that nested \emph{cups} still behave like a \emph{cup}, while a \emph{cup} and \emph{fork} may not.
Another approach would be to simulate the multiple objects involved, but that requires high-fidelity object models and physics interactions in simulation.

Nesting objects \IN{} and stacking them \ON{} one another exemplify two simple actions.
Object manipulation in industrial~\cite{kyrarini:auro19} or home furniture assembly~\cite{suarez:icra16,knepper:icra13} involves additional relationships, like \emph{fastening} one object to another (e.g., by tightening a screw or hammering a nail).
We provide a starting point for detecting when manipulation leads to containment, and hypothesize that prior information about the objects themselves will improve success detection accuracy for other, nuanced physical containment relationships.

\section{Conclusion~}
\label{sec:conclusions}

We provide a means by which a robot equipped with an egocentric, manipulator-mounted RGB-D camera can determine whether object placement actions result in one object being placed \IN{} or \ON{} another.
We propose a model solely trained on a robot's observations of outcomes, and then we show that task accuracy can be improved using prior visual and linguistic information about the objects involved.
We find that pretraining with crowdsourced annotations of whether objects can be placed \IN{} or \ON{} one another bootstraps useful feature extraction from this prior information.

\section*{Acknowledgements~}
This work was funded by the National Institute of Health (\#R01EB019335), NSF (CPS \#1544797, NRI \#1637748, \#IIS-1524371 \& \#1703166), the Office of Naval Research, the RCTA, Amazon, Honda, and DARPA's CwC program through ARO (W911NF-15-1-0543).

\IEEEtriggeratref{53}
\bibliography{references}
\bibliographystyle{IEEEtran}

\clearpage
\setcounter{table}{0}
\renewcommand{\thetable}{A\arabic{table}}
\section*{Appendix~}
\label{sec:appendix}

\subsection{Model Ablations}
To examine the contribution of each modality among egocentric camera images, object language referring expressions, and object static images, we perform a unimodal ablation analysis~\cite{thomason:naacl19}.
Under this ablation framework, the model architecture under examination remains unchanged, but ablated input modalities are set to zero tensors (\zv) at both training and inference time.

\begin{table}[t]
\caption{Object-only model ablations for the \textbf{All Pairs} task.}
\centering
\begin{tabular}{l c c r r}
	\toprule
    & \multicolumn{2}{c}{\bf Model ($M$)} & \multicolumn{2}{c}{\bf Prediction Correct $\uparrow$} \\
	& \multicolumn{2}{c}{Object Data} & & \\
	& Lang & Vis & \multicolumn{1}{c}{\IN{}} & \multicolumn{1}{c}{\ON{}} \\
	\midrule
	\multirow{5}{*}{\rotatebox{90}{\emph{Dev Fold}}}
	& \checkmark & \zv & $.86\pm.02$ & $.76\pm.01$ \\
	& \zv & \checkmark & $.94\pm.01$ & $.79\pm.01$ \\
	& \checkmark & \checkmark & $.86\pm.04$ & $.78\pm.01$ \\
	\cmidrule{2-5}
	& \multicolumn{2}{l}{Baseline (MC)} & $.87\pm.00$ & $.73\pm.00$ \\
	& \multicolumn{2}{l}{Baseline (Rand)} & $.49\pm.07$ & $.50\pm.03$ \\
	\midrule
	\multirow{5}{*}{\rotatebox{90}{\emph{Test Fold}}} 
	& \checkmark & \zv & $.86\pm.01$ & $.83\pm.01$ \\
	& \zv & \checkmark & $.88\pm.02$ & $.82\pm.01$ \\
	& \checkmark & \checkmark & $.87\pm.02$ & $.83\pm.01$ \\
	\cmidrule{2-5}
    & \multicolumn{2}{l}{Baseline (MC)} & $.84\pm.00$ & $.83\pm.00$ \\
    & \multicolumn{2}{l}{Baseline (Rand)} & $.51\pm.06$ & $.51\pm.03$ \\
	\bottomrule
	\multicolumn{5}{p{5cm}}{\checkmark indicates signal was included.}
\end{tabular}
\label{tab:lv_results_allpairs}
\end{table}

Performance of the \obj{} models on the auxiliary task of predicting Mechanical Turk workers' annotations for whether objects can be stacked \IN{} or \ON{} one another given object images and referring expressions is given in \tabref{tab:lv_results_allpairs} for the \emph{Test} and \emph{Development} data folds on {\bf All Pairs}.
In most cases, the prediction model using both language and vision signals achieves matching or higher accuracy than majority class, with the exception of losing about 1\% on the \emph{Development} fold on the \IN{} annotation prediction task.
The language-only and vision-only ablations reveal that either modality is competitive for predicting \IN{} and \ON{} annotations, with vision-only having the edge in most cases.

This auxiliary task facilitates pretraining the \egoPobj{} model based on the \egoobj{} architecture.
\tabref{tab:lv_results_robopairs} examines the performance of these models when ablating all three of egocentric, object language, and object vision modalities for detecting the results of actions on the {\bf Robot Pairs}.
Note that models with no egocentric input are essentially \emph{predicting} what happens to pairs of objects, as opposed to \emph{detecting} it from available egocentric scene information.

There are three high-level takeaways from these ablations.
First, In almost all cases, pretraining improves detection accuracy.
Where performance does not improve, it remains the same.
Second, performance on the {\it prediction} task for \IN{}, that is, when ablating the egocentric modality out, is competitive with the {\it detection} task of actually surveying the scene.
For the more complex \ON{} relationship, which can take different forms (as discussed in \sref{sec:analysis}), using the egocentric camera information improves performance over object priors alone.
Finally, as in the {\bf All Pairs} prediction experiments, both the language-only and vision-only ablations of object data perform well, suggesting that either alone may be enough to augment a detection model if the other is unavailable at inference time in practice.

\begin{table}[t]
\caption{\egoobj{} and \egoPobj{} model ablations on \textbf{Robot Pairs}.}
\centering
\begin{tabular}{l c c c r r}
	\toprule
    & \multicolumn{3}{c}{\bf Model ($M$)} & \multicolumn{2}{c}{\bf Detection Correct $\uparrow$} \\
	& \multirow{2}{*}{Ego} & \multicolumn{2}{c}{Object Data} & \\
	& & Lang & Vis & \multicolumn{1}{c}{\IN{}} & \multicolumn{1}{c}{\ON{}} \\
	\midrule
	\multirow{14}{*}{\rotatebox{90}{\emph{Dev Fold}}}
	& \zv & \checkmark & \zv & $.70\pm.03$ & $.56\pm.10$ \\
	& \zv & pre & \zv & $.72\pm.04$ & $.57\pm.09$ \\
	& \zv & \zv & \checkmark & $.71\pm.08$ & $.50\pm.06$ \\
	& \zv & \zv & pre & $.72\pm.07$ & $.53\pm.05$ \\
	& \zv & \checkmark & \checkmark & $.76\pm.08$ & $.58\pm.05$ \\
	& \zv & pre & pre & $.78\pm.08$ & $.60\pm.04$ \\
	\cmidrule{2-6}
	& \checkmark & \checkmark & \zv & $.67\pm.08$ & $.60\pm.08$ \\
	& \checkmark & pre & \zv & $.68\pm.08$ & $.62\pm.08$ \\
	& \checkmark & \zv & \checkmark & $.70\pm.10$ & $.58\pm.11$ \\
	& \checkmark & \zv & pre & $.72\pm.08$ & $.59\pm.13$ \\
	& \checkmark & \checkmark & \checkmark & $.70\pm.09$ & $.59\pm.07$ \\
	& \checkmark & pre & pre & $.73\pm.09$ & $.62\pm.07$ \\
	\cmidrule{2-6}
	& \multicolumn{3}{l}{Baseline (MC)} & $.32\pm.00$ & $.36\pm.00$ \\
	& \multicolumn{3}{l}{Baseline (Rand)} & $.49\pm.06$ & $.50\pm.06$ \\
	\midrule
	\multirow{14}{*}{\rotatebox{90}{\emph{Test Fold}}} 
	& \zv & \checkmark & \zv & $.79\pm.02$ & $.45\pm.05$ \\
	& \zv & pre & \zv & $.79\pm.02$ & $.48\pm.07$ \\
	& \zv & \zv & \checkmark & $.80\pm.04$ & $.46\pm.09$ \\
	& \zv & \zv & pre & $.81\pm.04$ & $.48\pm.06$ \\
	& \zv & \checkmark & \checkmark & $.80\pm.03$ & $.55\pm.04$ \\
	& \zv & pre & pre & $.79\pm.04$ & $.55\pm.04$ \\
	\cmidrule{2-6}
	& \checkmark & \checkmark & \zv & $.75\pm.06$ & $.54\pm.10$ \\
	& \checkmark & pre & \zv & $.80\pm.02$ & $.57\pm.07$ \\
	& \checkmark & \zv & \checkmark & $.75\pm.11$ & $.57\pm.10$ \\
	& \checkmark & \zv & pre & $.80\pm.05$ & $.56\pm.10$ \\
	& \checkmark & \checkmark & \checkmark & $.74\pm.07$ & $.59\pm.08$ \\
	& \checkmark & pre & pre & $.77\pm.05$ & $.59\pm.06$ \\
	\cmidrule{2-6}
    & \multicolumn{3}{l}{Baseline (MC)} & $.20\pm.00$ & $.32\pm.00$ \\
    & \multicolumn{3}{l}{Baseline (Rand)} & $.52\pm.05$ & $.51\pm.07$ \\
	\bottomrule
	\multicolumn{6}{p{7cm}}{\checkmark indicates signal was included, while ``pre" indicates models with object features pretrained from {\bf All Pairs} data.}
\end{tabular}
\label{tab:lv_results_robopairs}
\end{table}

\subsection{YCB Object Details}
We make a number of small changes and omissions from the full YCB Object Set when establishing our included objects $Y$ (Section~\ref{sec:task_and_data}).
In particular, we:
\begin{itemize}
\item exclude objects lacking camera image data used as vision information to the augmented model (\figref{fig:joint_model});
\item exclude \texttt{072-*\_toy\_airplane} parts \texttt{b-k};
\item do not split the two similar, \textit{medium-most-sized} \texttt{063-f\_cups} and \texttt{063-e\_cups} objects into different folds, to evaluate more conservatively; and
\item for \texttt{063-j\_cups}, we sometimes use a same-sized cup that is light blue instead of yellow during robot trials. 
\end{itemize}

\newpage

\subsection{Language Analysis}
\begin{figure}[t]
    \centering
    \includegraphics[width=\linewidth]{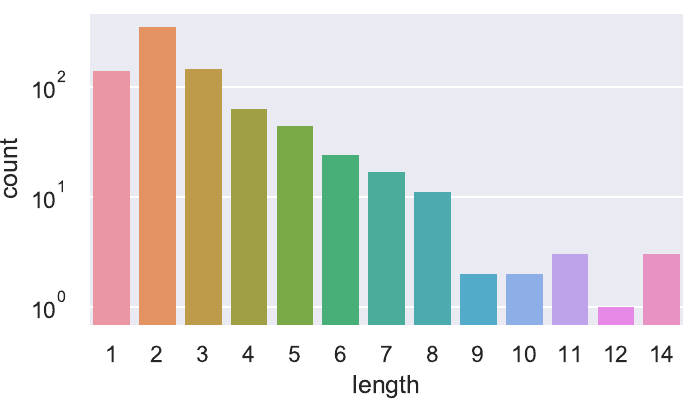}
    \vspace{-10pt}
    \caption{Distribution of language description lengths.}
    \label{fig:lengths}
\end{figure}

\begin{figure}[t]
    \centering
    \includegraphics[width=\linewidth]{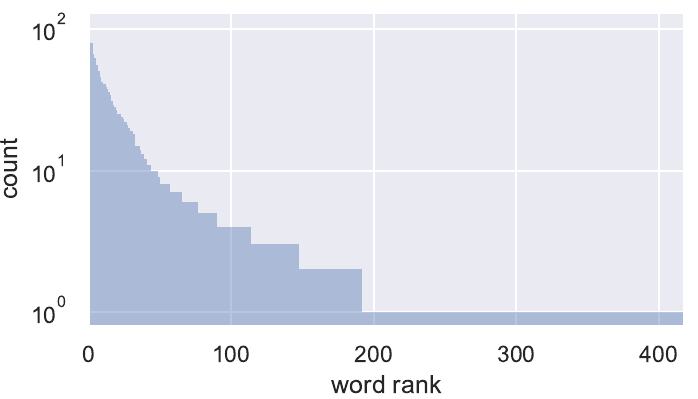}
    \vspace{-10pt}
    \caption{Frequency count for the $n$th ranked word in our descriptions.}
    \label{fig:frequencies}
\end{figure}

Our language annotations range up to 14 words in length (Figure \ref{fig:lengths}) and the importance of using pretrained vectors becomes clear as we see a large number of rare words (matching the expected Zipfian distribution of lexical tokens) in Figure \ref{fig:frequencies}.  First we analyze the description lengths and find that most responses contain fewer than five words.

Short sentences can take the form of a single entity (e.g. \textit{``apple"}) or have a single adjective (e.g. \textit{``blue cylinder"}).  In contrast, long descriptions may deviate from descriptions of form to function:
\begin{quote}
    \it thing you use to see how many spaces you move during a board game
\end{quote}
Understanding this requires much richer textual representations about the world.  Relatedly, precise descriptions which avoid using the name of an object also result in long descriptions.   Here we have a description of physical characteristics of the same die: 
\begin{quote}
\textit{white cube with different number of black dots on each side}.
\end{quote}

A similar pattern emerges when looking at the most and least frequent lexical types.  The most common are common adjectives and types:
\begin{quote}
    \textit{black, red, blue, yellow, cup, ball, box, plastic, ...}
\end{quote}
while the least common contain nuanced textures, world references, typos and counting:
\begin{quote}
    \textit{bumps, bloch, fleshy, eight, cleaning, avocado, ...}
\end{quote}

In this work, we use GloVe vectors as our pretrained world representations.  We leave the question of building better language representations that account for properties of the physical world to future work \cite{daruna19}.

\newpage
\subsection{Annotation Details}
For \ON{} and \IN{} labels across the five trials gathered for each \emph{Robot Pair}, the outcome label was annotated in \{\emph{Yes}, \emph{No}, \emph{Maybe}\}.
The \emph{Maybe} annotation denotes that there was mixed success across the five trials.
Similarly, for the Mechanical Turk annotations used during the auxiliary task for data augmentation, when annotators disagreed about the annotation in \{\emph{Yes}, \emph{No}\}, \emph{Maybe} was assigned.
Throughout our experiments, we round \emph{Maybe} annotations to the \emph{No} label for both robot trial and Mechanical Turk annotations.

For Mechanical Turk, annotators answer ``yes'', ``no'', or ``yes, but only if object A is rotated.''
We rounded this final option down to ``no'' for our task, but the original label may be useful for other manipulation tasks.

\end{document}